\documentclass[11pt]{article}
\usepackage{acl2016}
\usepackage{times}
\usepackage{latexsym}
\usepackage{color}
\usepackage{graphicx}
\usepackage{url}
\usepackage{amsfonts}
\usepackage[T1]{fontenc} 
\usepackage[utf8]{inputenc}
\aclfinalcopy

\title{Hierarchical Character-Word Models for Language Identification}

\author{Aaron Jaech$^{1}$ ~ George Mulcaire$^{2}$ ~ Shobhit Hathi$^{2}$ ~ Mari Ostendorf$^{1}$ ~ Noah A. Smith$^{2}$\\
$^1$Electrical Engineering \\
$^{2}$Computer Science \& Engineering \\
University of Washington, Seattle, WA, USA\\
{\tt ajaech@uw.edu, gmulc@uw.edu, shathi@uw.edu} \\ {\tt ostendor@uw.edu, nasmith@cs.washington.edu}
}

\date{}

\begin{document}

\maketitle

\begin{abstract}
Social media messages' brevity and unconventional spelling pose a
challenge
to language identification.  We introduce a hierarchical model that
learns character and contextualized word-level representations for language
identification.  Our method performs well against strong baselines,
and can also reveal code-switching.
% Language identification systems suffer when working with short
% texts or in domains with unconventional spelling, such as Twitter or other
% social media. Heirarchical character and word level models do well at language
% identification in the social media domain because they learn the morphological
% characteristics of each language and are also able to interpret each word in
% context to make a sentence level classification decision. We evaluate our
% hybrid model on TweetLID, a dataset of six Iberian languages, and on another
% 70 language Twitter dataset.
\end{abstract}

\section{Introduction}

Language identification (language ID), despite being described as a solved problem more than ten years ago \cite{mcnamee2005}, remains a difficult problem. 
Particularly when working with short texts, informal styles, or closely related language pairs, it is an active area of research \cite{Gella2014YW,Wang2015ALD,baldwin2010language}. 
These difficult cases are often found in social media content.
Progress on language ID in social media is needed especially since all downstream tasks, like machine translation or semantic parsing, depend on correct language ID.

This paper brings continuous representations for language data, which have produced new states of the art for language modeling \cite{mikolov2010recurrent}, machine translation \cite{bahdanau2014neural}, and other tasks, to language ID.
We adapt a hierarchical character-word neural architecture from 
\newcite{kim2015character}, demonstrating that it works well for
language ID.
Our model, which we call C2V2L (``character to vector to language'')
is hierarchical in the sense that it explicitly builds a continuous representation
for each word from its character sequence, capturing orthographic and
morphology-related patterns, and then combines those word
level representations in context, finally classifying the full word sequence.
Our model does not require any special handling of casing or punctuation nor do we need to remove the URLs, usernames or hashtags, and it is trained end-to-end using standard
procedures.

We demonstrate the model's state-of-the-art performance in experiments
on two difficult language ID datasets consisting of tweets.
This hierarchical technique works better than previous classifiers using character or word
$n$-gram features as well as a similar neural model that treats an entire
tweet as a single character sequence. 
We find further that the model can benefit from additional out-of-domain data, unlike much previous work, and with very little modification can annotate word-level code-switching.  
We also confirm that smoothed character n-gram language models perform very well for language ID tasks.

\section{Model}

Our model has two main components, though they are trained together, end-to-end.\footnote{Code will be made
  available on GitHub after publication.} 
The first, ``char2vec,'' applies a convolutional neural network (CNN)
to a whitespace-delimited word's Unicode character sequence, providing
a word vector.  The second is a bidirectional LSTM recurrent neural network (RNN) that
maps a sequence of such word vectors to a label (a language).

\subsection{Char2vec}

The first layer of char2vec embeds characters.  An embedding is
learned for each Unicode code point that appears at least twice in the training data, 
including punctuation, emoji, and other symbols. 
If $C$ is the set of characters then we let the size of the character embedding
layer be $d = \lceil \log_2 |C| \rceil$. (If each dimension of the character embedding vector holds 
just one bit of information then $d$ bits should be enough to uniquely encode each character.)
The character embedding matrix is $\mathbf{Q} \in \mathbb{R}^{d \times |C|}$.
Words are given to the model as a sequence of characters. When each character in a word of length $l$
is replaced by its embedding vector we get a matrix $\mathbf{C} \in \mathbb{R}^{d \times (l + 2)}$. There are $l + 2$
columns in $C$ because padding characters are added to the left and right of each word.

The char2vec architecture uses two sets of filter banks. The first set is comprised of matrices
$\mathbf{H}_{a_i} \in \mathbb{R}^{d \times 3}$ where $i$ ranges from $1$ to $n_1$. The matrix $\mathbf{C}$ is narrowly convolved
with each of the $\mathbf{H}_{a_i}$,  a bias term $b_a$ is added and an ReLU non-linearity, $\textrm{ReLU}(x) = \max(0, x)$, 
is applied to produce an output 
$\mathbf{T}_1 = \textrm{ReLU}(\textrm{conv}(\mathbf{C}, \mathbf{H}_a) + \mathbf{b}_a)$. 
$\mathbf{T}_1$ is of size $n_1 \times l$ with one row for each of the filters and one column for each of the characters in the input word. 
% size is now reduced by 2 because the left and right padding symbols go away after the first convolution
Since each of the $\mathbf{H}_{a_i}$ is a filter with a width of three characters, the columns of
$\mathbf{T}_1$ each hold a representation of a character tri-gram. During training, we apply dropout on $\mathbf{T}_1$ to 
regularize the model.
The matrix $\mathbf{T}_1$ is then convolved with a second set of filters $\mathbf{H}_{b_i} \in \mathbb{R}^{n_1 \times w}$
where $b_i$ ranges from 1 to $3 n_2$ and $n_2$ controls the number of filters of each of the possible widths,
$w = 3, 4$, or $5$. Another convolution and ReLU non-linearity is applied to get $\mathbf{T}_2 = \textrm{ReLU}(\textrm{conv}(
 \mathbf{T}_1, \mathbf{H}_b) + \mathbf{b}_b)$. Max-pooling across time is used to create a fix-sized vector $\mathbf{y}$ from $\mathbf{T}_2$.
The dimension of $\mathbf{y}$ is $3 n_2$, corresponding to the number of filters used.

Similar to \newcite{kim2015character} who use a highway network after the max-pooling layer, we apply a residual network layer.
Both highway and residual network layers allow values from the previous layer to pass through unchanged but the residual layer is preferred in our case because it uses half as many parameters \cite{he2015deep}.
The residual network uses a matrix 
$\mathbf{W} \in \mathbb{R}^{3 n_2 \times 3 n_2}$ 
and bias vector 
$\mathbf{b}_3$ 
to create the vector 
$\mathbf{z} = \mathbf{y} + f_R(\mathbf{y})$
where $f_R(\mathbf{y})=\textrm{ReLU}(\mathbf{W} \mathbf{y} + \mathbf{b}_3)$. 
The resulting vector 
$\mathbf{z}$ 
is used as a word embedding vector in the word-level LSTM portion of the model.

There are three differences between our version of the model and the one described by \newcite{kim2015character}. 
First, we use two layers of convolution instead of just one, inspired by \cite{ling2015finding} which uses a 2-layer LSTM for character modeling. 
Second, we use the ReLU function as a nonlinearity as opposed to the tanh function. 
ReLU has been highly successful in computer vision applications in conjunction with convolutional layers
\cite{jarrett2009best}. 
Finally, we use a residual network layer instead of a highway network layer after the max-pooling
step, to reduce the model size.

\begin{figure}[h]
\centering
\includegraphics[width=0.4\textwidth]{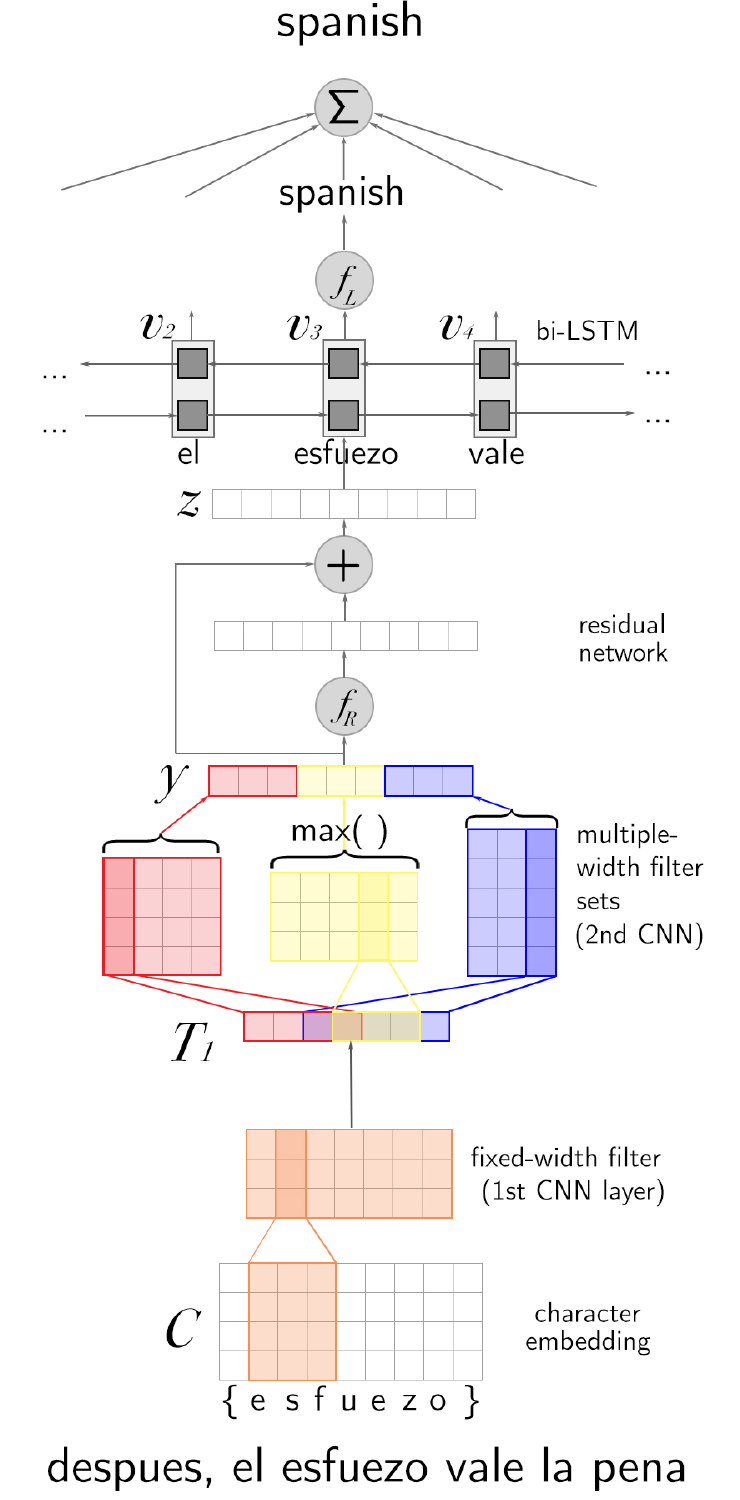}
\caption{C2V2L model architecture. The model takes the word ``esfuezo,'' a misspelling of the Spanish word ``esfuerzo,'' and maps it to a word vector via the two CNN layers and the residual layer. The word vector is then combined with others via the LSTM, and the words' language predictions averaged to get a tweet prediction. 
%\georgecomment{@Noah---suggestions for figure?}
}
\label{fig:arch}
\end{figure}

It is possible to use bi-LSTMs instead of convolutional layers in char2vec as done by \newcite{ling2015finding}.
We explored this option in preliminary experiments and found that using convolutional layers has several advantages, including a 
large improvement in speed for both the forward and backward pass, many fewer
parameters, and improved language ID accuracy.
 
\subsection{Sentence-level Language ID}
The sequence of word embedding vectors is processed by a bi-LSTM, which outputs a sequence of
vectors, $ [ \mathbf{v}_1, \mathbf{v}_2, \mathbf{v}_3 \dots \mathbf{v}_T]$ where $T$ is the number of
words in the tweet. All LSTM gates are used as defined by \newcite{sak2014long}.
Dropout is used as a regularizer on the inputs to the LSTM \cite{zaremba2014recurrent}. The output vectors $\mathbf{v}_i$ are transformed into probability distributions over the set of languages
by applying an affine transformation followed by a softmax:
\[  \mathbf{p}_i  = f_L(\mathbf{v}_i) = \frac{\exp( \mathbf{A} \mathbf{v}_i + b)}{\sum_{t=1}^T{\exp( \mathbf{A} \mathbf{v}_t + b)}}\] 
(These word-level predictions, we will see in
\S\ref{sec:codeswitching}, are useful for annotating code-switching.) The sentence-level prediction $\mathbf{p}_S$ is then given by an average of the word-level language 
predictions:
\[  \mathbf{p}_S = \frac{1}{T} \sum_{i=1}^T \mathbf{p}_i \]

The final affine transformation can be interpreted as a language
embedding, where each language is represented by a vector of the same
dimensionality as the LSTM outputs. The goal of the LSTM then is (roughly) to maximize the dot product of each word's representation with the language embedding(s) for that sentence.
The only supervision in the model comes from computing the loss of sentence-level predictions.

\section{Tasks and Datasets}

We consider two datasets: TweetLID and Twitter70. Summary statistics for each of the datasets are
provided in Table \ref{table:datasets} including the number of training examples per dataset.

\subsection{TweetLID} \label{sec:tweetlid}
The TweetLID dataset \cite{zubiaga2014overview} comes from a language ID shared task that focused on six commonly spoken languages of the Iberian peninsula: Spanish, Portuguese, Catalan, Galician, English, and Basque. 
There are approximately 15,000 Tweets in the training data and 25,000 in the test set. 
The data is unbalanced, with the majority of examples being in the Spanish language. 
The ``undetermined'' label ('und'), comprising 1.4\% of the training data, is used for Tweets
that use only non-linguistic tokens or belong to an outside language.
Additionally, some Tweets are ambiguous ('amb') among a set of languages (2.3\%), or code-switch between languages (2.4\%). 
The evaluation criteria take into account all of these factors, requiring prediction of at least one acceptable language for an ambiguous Tweet or all languages present for a code-switched Tweet. 
The fact that hundreds of Tweets were labeled ambiguous or undetermined by annotators who were native speakers of these languages reveals the difficulty of this task.

For tweets labeled as ambiguous or containing multiple languages, 
the training objective distributes the ``true'' probability mass
evenly across each of the label languages, 
e.g. 50\% Spanish and 50\% Catalan. 

The TweetLID shared task had two tracks: one that restricted participants to only use the 
official training data and another that was unconstrained, allowing the use of any external data. 
There were 12 submissions in the constrained track and 9 in the
unconstrained track. Perhaps 
surprisingly, most participants performed worse on the unconstrained track than they did on 
the constrained one.

As supplementary data for our unconstrained-track experiments, we collected data 
from Wikipedia for each of the six languages in the TweetLID corpus. Participants in 
the TweetLID shared task also used Wikipedia as a data source for the unconstrained track. 
We split the text into 25,000 sentence fragments per language, with each fragment of length
comparable to that of a Tweet. 
The Wikipedia sentence fragments are easily distinguished from tweets. Wikipedia fragments are more formal and are more likely to use complex words; for example, one fragment reads ``ring homomorphisms are identical to monomorphisms in the category of rings.'' In contrast, tweets tend to use variable spelling and more simple words, as in ``Haaaaallelujaaaaah http://t.co/axwzUNXk06'' and ``@justinbieber: Love you mommy http://t.co/xEGAxBl6Cc http://t.co/749s6XKkgK awe $\heartsuit$''.
Previous work confirms that language ID is more 
challenging on social media text than sentence fragments taken from more formal text,
like Wikipedia \cite{carter2012exploration}. Despite the domain
mismatch, we find in \S\ref{sec:unconstrained} that additional text at training time is useful for our model.

The TweetLID training data is too small to divide into training and validation sets. 
We created a tuning set by adding samples taken from Twitter70 and from the 2014 Workshop on Computational Approaches to Code Switching \cite{solorio2014overview} to the official TweetLID training data.
We used this augmented datset with a 4:1 train/eval split for hyperparameter tuning.\footnote{We used this augmented data to tune hyperparameters for both constrained and unconstrained models. However, after setting hyperparameters, we trained our constrained model using only the official training data, and the unconstrained model using only the training data + Wikipedia. Thus, no extra data was used to learn actual model parameters for the constrained case.}

\begin{table}[]
\centering
\begin{tabular}{c|cc}
 & \textbf{TweetLID} & \textbf{Twitter70} \\ \hline
Tweets & 14,991 & 58,182 \\
Character vocab & 956 & 5,796 \\
Languages & 6 & 70 \\
Code-switching? & Yes & Not Labeled \\
Balanced? & No & Roughly % not exactly balanced but much more balanced then tweetlid
\end{tabular}
\caption{Dataset characteristics.
\label{table:datasets}}
\end{table}

\subsection{Twitter70}
The Twitter70 dataset was published by the Twitter Language Engineering Team in 
November 2015.\footnote{For clarity, we refer to this data as
  ``Twitter70'' but it can be found in the Twitter blog post under the
  name ``recall oriented.'' See \url{http://t.co/EOVqA0t79j}}
% full url http://blog.twitter.com/2015/evaluating-language-identification-performance}} 
The languages come from the Afroasiatic, Dravidian, Indo-European, 
Sino-Tibetan, and Tai-Kadai families. Each person who wants to use the data must 
redownload the Tweets using the Twitter API. In between the time when the 
data was published and when it is downloaded, some of the Tweets can be lost 
due to account deletion or changes in privacy settings. At the time when the
data was published there were approximately 1,500 Tweets for each language.
We were able to download 82\% of the Tweets but the amount that we could 
access varied by language with as many as 1,569 examples for Sindhi and 
as few as 371 and 39 examples for Uyghur and Oriya, respectively. The 
median number of Tweets per language was 1,083. To our knowledge, there
are no published benchmarks on this dataset.

Unlike TweetLID, the Twitter70 data has no unknown or ambiguous labels. Some tweets
do contain code-switching but that is not labeled as such; a single language is assigned. 
There is no predefined test set so we used the last digit of the identification number to partition them. 
Identifiers ending in zero (15\%) were used for the test set and those ending in one (5\%) were 
used for tuning.

When processing the input at the character level, the vocabulary for each data source
is defined as the set of Unicode code-points that occur at least twice in the training
data: 956 and 5,796 characters for TweetLID and Twitter70, respectively. 
A small number of languages, such as Mandarin, are responsible for most of the characters in the Twitter70 vocabulary.

One recent work processed the input one byte at a time instead of by character
\cite{gillick2015multilingual}. In early experiments, we found that when using bytes 
the model would often make mistakes that should have been much more obvious
from the orthography alone. We do not recommend using the byte sequence for language ID.

\section{Implementation Details}

\subsection{Preprocessing}   \label{sec:preprocessing}
An advantage of the hybrid character-word model is that only limited preprocessing is required. 
The runtime of training char2vec is proportional to the longest word in a minibatch. 
The data contains many long and repetitive character sequences such as ``hahahaha...'' or ``arghhhhh...''. 
To deal with these, we restricted any sequence of repeating characters to at most five repetitions where the repeating pattern can be from one to four characters. 
There are many tweets that string together large numbers of twitter username or hashtags without spaces between them. 
These create extra long ``words'' that cause our implementation to use more memory and do extra computation during training. 
To solve this we enforce the constraint that there must be a space before any URL, username, or hashtag. 
To deal with the few remaining extra-long character sequences, we force word breaks in non-space character sequences every 40 bytes. 
This primarily affects languages that are not space-delimited like Chinese. We do not perform any special handling of casing or punctuation nor do we need to remove the URLs, usernames or hashtags as has been done in previous work \cite{zubiaga2014overview}. The same preprocessing is used when training the $n$-gram models.

\subsection{Training and Tuning}

Training is done using minibatches of size 25 and a learning rate of 0.001 using the Adam method for 
optimization \cite{kingma2014adam}. 
For the Twitter70 dataset we used 5\% held out data for tuning and 15\% for evaluation. 
To tune, we trained 15 models with random hyperparameters and selected the one that performed the best on the development set. 
Training is done for 80,000 minibatches for TweetLID and 100,000 for the Twitter70 dataset. 

There are only four hyperparameters to tune for each model: the number of filters in the first convolutional layer, the number of filters in the second convolutional layer, the size of the word-level LSTM vector, and the dropout rate.
We designed our model to have as few hyperparameters as possible so that we could be more confident that our models were properly tuned. 
The selected hyperparameter values are listed in Table \ref{table:hyperparams}.
There are 193K parameters in the TweetLID model and 427K in the Twitter70 model.

\begin{table}[]
\centering
\begin{tabular}{ccc}
  \textbf{Parameter}   & \textbf{TweetLID} & \textbf{Twitter70} \\ \hline
1st Conv. Layer ($n_1$) &         50          & 59                  \\
2nd Conv. Layer ($n_2$) &       93            & 108                 \\
LSTM    &      23              & 38                  \\
Dropout  &       25\%        & 30\%              \\
Total Params. &   193K  &    346K                   \\
\end{tabular}
\caption{Hyperparameter settings for selected models.}
\label{table:hyperparams}
\end{table}

\section{Experiments} 

%First, we compare to several existing systems: the winning submissions for the two conditions in the 
%original TweetLID shared task, and \texttt{langid.py}, a popular
%open-source language ID package, on the TweetLID dataset. Next we compare against  \texttt{langid.py} and our $n$-gram baseline on Twitter70, and finally we test our model's word level performance on a code-switching dataset.

For all the studies below on language identification, we compare to two baselines: i) \texttt{langid.py}, a popular open-source language ID package, and ii) a classifier using $n$-gram character language models. For the TweetLID dataset, additional comparisons are included as described next. In addition, we test our model's word level performance on a code-switching dataset.

The first baseline, based on the \texttt{langid.py} package,
uses a naive Bayes classifier over \textit{byte} $n$-gram features
\cite{lui2012langid}. The pretrained model distributed with the
package is designed to perform well on a wide range of domains, and
achieved high performance on ``microblog messages'' (tweets) in the
original paper. \texttt{langid.py} uses feature selection for domain
adaptation and to reduce the model size; thus, retraining it on
in-domain data as we do in this paper does not provide an entirely
fair comparison. However, we include it for its popularity and
importance.

The second baseline is built from character $n$-gram language models
for each language $\ell$.  It assigns each tweet according to
$\ell^*=\arg\max_{\ell} p(\textrm{tweet} \mid \ell)$, i.e., applying
Bayes' rule with a uniform class prior
\cite{dunning1994statistical}. 
For TweetLID, the rare 'und' was handled with a rejection model. 
Specifically, after $\ell^*$ is chosen, a log likelihood ratio test is applied to decide
whether to reject the decision in favor of the 'und' class, using the
language models for $\ell^*$ and 'und' with a threshold chosen to optimize $F_1$
on the dev set.  The models were trained using Witten-Bell smoothing
\cite{bell1989modeling}, but otherwise the default parameters of the
SRILM toolkit \cite{stolcke2002srilm} were used.\footnote{Witten-Bell
  smoothing works well when working with comparatively small
  vocabularies such as with character sets.}
TweetLID model training
ignores tweets labeled as ambiguous or containing multiple languages,
and the unconstrained TweetLID models use a simple interpolation of TweetLID
and Wikipedia component models. The $n$-gram order was chosen to minimize perplexity using 5-fold cross validation, yielding $n=5$ for TweetLID and Twitter70, and
$n=6$ for Wikipedia.

Note that both of these baselines are generative, learning separate models for each language. In contrast, the neural network models explored here are trained on all languages, so parameters may be shared across languages. In particular, for the hierarchcial model, a character sequence corresponding to a word in more than one language (e.g. ``no'' in English and Portuguese) has a language-independent word embedding.

\subsection{TweetLID:  Constrained Track}

In the constrained track of the 2014 shared task, \newcite{hurtado2014elirf} attained the highest performance (75.2 macroaveraged $F_1$). 
They used a set of one-vs-all SVM classifiers with character $n$-gram features, and returned all languages for which the classification confidence was above a fixed
threshold.
This provides our third, strongest baseline.

In the unconstrained track, the winning team was \newcite{gamallo2014comparing}, using a naive Bayes classifier on word unigrams. 
They incorporated Wikipedia text to train their model, and were the only team in the competition whose unconstrained model outperformed their constrained one.
We compare to their constrained-track result here.

We also consider a version of our model, ``C2L,'' which uses
only the char2vec component of C2V2L, treating the entire tweet as a
single word.  This tests the value of the intermediate word
representations in C2V2L; C2L has no explicit word
representations.  Hyperparameter tuning was carried out separately for C2L.

\paragraph{Results} The first column of Table \ref{tab:tweetlid}
 shows the aggregate results across all labels.   Our model achieves the state of the art on 
this task, surpassing the shared task winner, \newcite{hurtado2014elirf}.
As expected, C2L fails to match the performance of C2V2L, demonstrating
that there is value in the hierarchical representations. The
performance of the $n$-gram LM baseline is notably strong, beating
eleven out of the twelve submissions to the TweetLID shared task.
% \begin{table}[t]
% \centering
% \begin{tabular}{l|r}
% \textbf{Model}  								& \textbf{$F_1$}   \\  \hline
% $n$-gram LM   								& 75.0  \\ 
% \texttt{langid.py}  							& 68.9  \\ 
% \newcite{hurtado2014elirf}    			& 75.2   \\
% \newcite{gamallo2014comparing}  	& 72.6 \\
% C2V2L 											& \textbf{76.2}    \\
% C2L 												& 72.7 \\
% \end{tabular}
% \caption{$F_1$ scores on the TweetLID language ID task (constrained track). The scores for \protect\newcite{hurtado2014elirf} and \protect\newcite{gamallo2014comparing} are as reported in \protect\newcite{zubiaga2014overview}. }
% \label{tab:tweetlid}
% \end{table}
We also report category-specific performance for our models and baselines in Table \ref{tab:tweetlid2}. 
Note that performance on underrepresented categories such as 'glg' and 'und' is much
lower than the other categories. The category breakdown is not available for previously published results. 

\begin{table*}[t]
\centering
\begin{tabular}{l|r||r|r|r|r|r|r|r|r}
\textbf{Model}  & \textbf{Avg.} $F_1$ & \textbf{eng}  & \textbf{spa}  & \textbf{cat}  & \textbf{eus}  & \textbf{por}  & \textbf{glg}  & \textbf{und} & \textbf{amb}  \\ \hline
$n$-gram LM    & 75.0	& 74.8              & 94.2            	& 82.7           	  &  74.8            	&  \textbf{93.4}	&  49.5             &  \textbf{38.9}	& 87.0 \\
\texttt{langid.py} & 68.9	& 65.9              &  92.0
                                                                      &  72.9             &  70.6            	& 89.8               	& 52.7              &  18.8             & 83.8  \\
\newcite{hurtado2014elirf} & 75.2 & &&&&&&& \\
\newcite{gamallo2014comparing} & 72.6 & &&&&&&& \\
C2L            & 72.7      	& 73.0              &  93.8            	&  82.6             &  75.7             &  89.4              	&  57.0             &  18.0           	& 92.1 \\
C2V2L             & \bf 76.2  	& \textbf{75.6}  & \textbf{94.7} 	&  \textbf{85.3} &  \textbf{82.7}	&  91.0             	& \textbf{58.5}  &  27.2           	& \textbf{94.5} \\
\end{tabular}
\caption{$F_1$ scores on the TweetLID language ID task (constrained
  track), averaged and per language category (including undetermined
  and ambiguous). The
scores for \protect\newcite{hurtado2014elirf} and
\protect\newcite{gamallo2014comparing} are as reported in
\protect\newcite{zubiaga2014overview}; per-language scores are not available.
\label{tab:tweetlid}\label{tab:tweetlid2}}
\end{table*}

One important advantage of our model is its ability to handle special categories
of tokens that would otherwise require special treatment as
out-of-vocabulary symbols, such as URLs, hashtags,
emojis, usernames, etc. Anecdotally, we observe that the input gates of the word-level
LSTM are less likely to open for these special classes of tokens. 
This is consistent with the hypothesis that the model has learned to
ignore tokens that are non-informative with respect to language ID.

\subsection{TweetLID:  Unconstrained Track}
\label{sec:unconstrained}

We augmented C2V2L's training data with 25,000 fragments of Wikipedia text, weighting the TweetLID training examples ten times more strongly.  
After training on the combined data, we ``fine-tune'' the model on the TweetLID data for 2,000 minibatches.
We found this stage necessary to correct for bias away from the undetermined language category, which does not occur in the Wikipedia data. 
The same hyperparameters were used as in the constrained experiment.

For the $n$-gram baseline, we interpolated the language models trained on TweetLID and Wikipedia for each language. 
Interpolation weights given to the Wikipedia language models, set by cross-validation, ranged from 16\% for Spanish to 39\% for Galician, the most and least common labels respectively.

We also compare to the unconstrained-track results of \newcite{hurtado2014elirf} and \newcite{gamallo2014comparing}.

\paragraph{Results} 
The results for these experiments are given in Table \ref{table:unconstrained}.
Like \newcite{gamallo2014comparing}, we see a benefit from the use of out-of-domain data, giving a new state of the art on this task as well.  
Overall the $n$-gram language model does not benefit from Wikipedia, but we observe that if the undetermined category, which is not observed in Wikipedia training data, is ignored, then there is a net gain in performance. 

\begin{table}[]
\centering
\begin{tabular}{l|r|r}
\textbf{Model}   & \textbf{$F_1$} & $\Delta$\\ \hline
\newcite{hurtado2014elirf}    			& 69.7     	&  --4.5    \\
\newcite{gamallo2014comparing}   	& 75.3     	&  +2.7    \\
$n$-gram LM 								& 74.7     	&  --0.3   \\
C2V2L  											&\bf 77.1	&  +0.9 
\end{tabular}
\caption{$F_1$ scores for the unconstrained data track of the TweetLID
  language ID task. $\Delta$ measures change in absolute $F_1$ score from 
  the constrained condition.}
\label{table:unconstrained}
\end{table}

\begin{table*}[t]
\centering
\begin{tabular}{cl|cl|cl}
\multicolumn{2}{c}{\textbf{couldn't}} & \multicolumn{2}{c}{\textbf{@maria\_sanchez}} & \multicolumn{2}{c}{\textbf{noite}} \\ \hline
can't    & 0.84 & @Ainhooa\_Sanchez & 0.85 & Noite   & 0.99 \\
'don't   & 0.80 & @Ronal2Sanchez:   & 0.71 & noite.  & 0.98 \\
ain't    & 0.80 & @maria\_lsantos   & 0.68 & noite?  & 0.98 \\
don't    & 0.79 & @jordi\_sanchez   & 0.66 & noite.. & 0.96 \\
didn't   & 0.79 & @marialouca?      & 0.66 & noite,  & 0.95 \\
Can't    & 0.78 & @mariona\_g9      & 0.65 & noitee  & 0.92 \\
first    & 0.77 & @mario\_casas\_   & 0.65 & noiteee & 0.90
\end{tabular}
\caption{Top seven most similar words from the training data and their cosine similarities for
inputs ``couldn't'', ``@maria\_sanchez'', and ``noite''.}
\label{table:closest_words}
\end{table*}

In Table \ref{table:closest_words}, we show the top seven neighbors to selected input words based on cosine similarity.
In the left column we see that words with similar features, such as the presence of ``n't'' contraction, can be grouped together by char2vec. 
In the middle column, an out-of-vocabulary username is supplied and similar usernames are retrieved. 
When working with $n$-gram features, removing usernames is common, but some previous 
work demonstrates that they still carry useful information for predicting the language of the tweet \cite{jaechyour}.
The third example,``noite'' (the Portuguese word for ``night''), shows that the word embeddings are largely invariant to changes in punctuation and capitalization.

\subsection{Twitter70}

We compare C2V2L to \texttt{langid.py} and the 5-gram language model
on the Twitter70 dataset; see Table~\ref{tab:twitter}. Since this data has not been published on,
these are the only two comparisons we have. Although the 5-gram model achieves the best performance, the results are virtually identical to those for C2V2L except for the closely-related Bosnian-Croatian language pair.

\begin{table}[]
\centering
\begin{tabular}{r|r|r}
\textbf{Model}  & \textbf{$F_1$} \\  \hline
\texttt{langid.py}    	& 87.9  \\ 
5-gram LM  	& 93.8  \\ 
C2V2L   (ours)  	& 91.2 \\  
\end{tabular}
\caption{$F_1$ scores on the Twitter70 dataset.}
\label{tab:twitter}
\end{table}

The lowest performance for all the models is on closely related language pairs.
For example using the C2V2L model, the $F_1$ score for Danish is only 62.7 due to confusion with the mutually 
intelligble Norwegian \cite{van2008linguistic}. Distinguishing Bosnian from 
Croatian, two varieties of a single language, is also difficult. Languages that have unique 
orthographies such as Greek and Korean are identified with near perfect accuracy by each of the models.

\begin{figure}[h]
\centering
\includegraphics[width=0.49\textwidth]{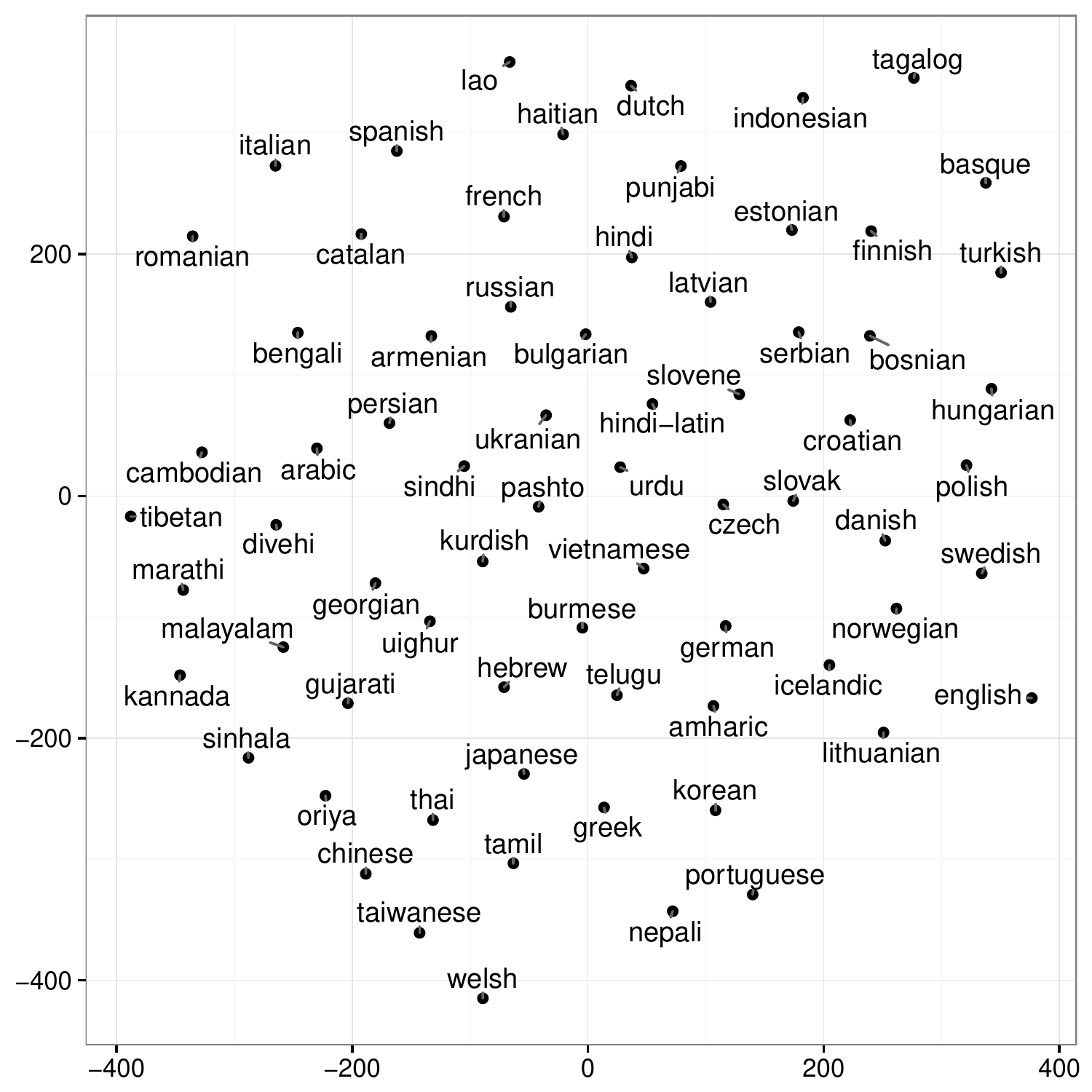}
\caption{T-SNE plot of language embedding vectors.}
\label{fig:lang_embed}
\end{figure}

A potential advantage of the C2V2L model over the $n$-gram models is the ability to share information
between related languages. In Figure \ref{fig:lang_embed} we show a T-SNE plot of the language embedding
vectors taken from the softmax layer of our model trained with a rank constraint of 10 on the softmax layer.
Many of the languages in the plot appear close to related languages.\footnote{The rank constraint was added for visualization; without it, the model learns language embeddings which are all roughly orthogonal to each other, making T-SNE visualization difficult.}
Another advantage of the C2V2L model is that the word-level predictions provide an indication of code-switching, explored next.

\subsection{Code-Switching} \label{sec:codeswitching}
Because C2V2L produces language predictions for every word before
making the tweet-level prediction, the same architecture can be used in word-level analysis of code-switched text, switching between multiple languages. 
Training a model that predicts code-switching at the token level requires a dataset that has language labels at this level. 
We used the Spanish-English dataset from the EMNLP 2014 shared task on Language Identitication in Code-Switched Data \cite{solorio2014overview}: a collection of monolingual and code-switched tweets in English and Spanish.

To train and predict at the word level, we simply remove the final average
over the word predictions, and calculate the loss as the
sum of the cross-entropy between each
word's prediction and the corresponding gold label. Both the char2vec
and word LSTM components of the model architecture are unaffected,
other than retraining their parameters.\footnote{Potentially, both sentence-level and word-level supervision could be used to train the same model, but we leave that for future work.} To tune hyperparameters, we trained 10 models with random parameter settings on 80\% of the data from the training set, and chose the settings from the model that performed best on the remaining 20\%. We then retrained on the full training set with these settings.

C2V2L performed well at this task, scoring 95.1 $F_1$ for English
(which would have achieved second place in the shared task, out of
eight entries),
94.1 for Spanish (second place), 36.2 for named entities (fourth place) and 94.2 for
Other (third place).\footnote{Full results for the 2014 shared task are omitted for space but can be found at \url{http://emnlp2014.org/workshops/CodeSwitch/results.php}. } 
While our code-switching results are not quite state-of-the-art, they show that our model learns to make accurate word-level predictions.

\section{Related Work}

The task of language ID has a long history both in the speech domain
\cite{house1977toward} and for text \cite{cavnar1994n}. Previous work on the text domain mostly 
uses word or character $n$-gram features combined with linear classifiers \cite{hurtado2014elirf,gamallo2014comparing}.

Recently published work by \newcite{xeroxTweetlid} showed that combining an $n$-gram language model classifier (similar to our $n$-gram baseline) with information from the Twitter social graph improves language ID on TweetLID from 74.7 to 76.6 $F_1$ which is only slightly better than our model's performance of 76.2.

\newcite{Bergsma2012Language} created their own multilingual Twitter dataset and tested both
a discriminative model based on $n$-grams plus hand-crafted features and a compression-based classifier.
Since the Twitter API requires researchers to re-download tweets based on their identifiers, published datasets quickly go out of date when the tweets in question are no longer available online, making it difficult to compare against prior work.

Several other studies have investigated the use of character sequence models in language processing. 
These techinques were first applied only to create word embeddings \cite{Santos+14,Santos+15} and then later extended to have the word embeddings feed directly into a word level RNN.
Applications include part-of-speech (POS) tagging \cite{Ling+15b}, language modeling \cite{ling2015finding}, dependency parsing \cite{Ballesteros+15},  translation \cite{Ling+15b}, and slot filling text analysis  \cite{Jaech+16}. The work is divided in terms of whether the character sequence is modeled with an LSTM or CNN, though virtually all now leverage the resulting word vectors in a word-level RNN. We are not aware of prior results comparing LSTMs and CNNs on a specific task, but the reduction in model size compared to word-only systems is reported to be much higher for LSTM architectures. All analyses report that the greatest improvements in performance from character sequence models are for infrequent and previously unseen words, as expected.

In addition to the 2014 Workshop on Computational Approaches to Code Switching \cite{solorio2014overview}, word-level language identification in code-switched text has been studied by \newcite{Mandal2015AdaptiveVI} in the context of question answering and by \newcite{Garrette2015UnsupervisedCF} in the context of language modeling for document transcription.
Both used primarily character $n$-gram features.
The use of character representations is well motivated for code-switching LID since the presence of multiple languages means that one is more likely to encounter a previously unseen word.

\section{Conclusion}

We present C2V2L, a hierarchical neural model for language ID that
outperforms previous work on the challenging TweetLID task.   
We also find that smoothed character $n$-gram language models can work well as
classifiers for language ID for short texts. 
Without feature engineering, our $n$-gram model beat eleven out of the twelve submissions in the TweetLID shared task, and gives the best reported performance on the Twitter70 dataset, where training data for some languages is quite small.
In future work, we plan to further adapt C2V2L for analyzing code-switching, having found that it already offers good 
%\georgecomment{Still unsure of word choice! Mari votes for ``quite good''} 
performance without any change to the architecture.

%Previous work on code-switching detection has typically used small
%datasets with only two languages \cite{rosner2007tagging}. More data
%is needed to make progress in this area. One promising direction is to use an initial model to bootstrap the collection of additional code-switched data.

%\section*{Acknowledgments}

\bibliography{emnlp2016}
\bibliographystyle{acl2016}

\end{document}